\documentclass[10pt,twocolumn,letterpaper]{article}

\usepackage[pagenumbers]{wacv} 

\definecolor{wacvblue}{rgb}{0.21,0.49,0.74}
\usepackage[pagebackref,breaklinks,colorlinks,allcolors=wacvblue]{hyperref}

\def\wacvPaperID{1226} 
\def\confName{WACV}
\def\confYear{2027}

\title{MYOPIA: Understanding the Limits of Tree-Based Reasoning in Vision-Language Models}

\author{
  Islam Mesabah$^{1}$ \quad
  Sary Elmansoury$^{1}$ \quad
  Gerrit Großmann$^{1}$ \\
  Peter Neigel$^{1}$ \quad
  Sebastian Vollmer$^{1}$\\
  $^{1}$German Research Centre for Artificial Intelligence (DFKI), Kaiserslautern, Germany\\
  $^{1}${\ttfamily\small \{first\_name.last\_name\}@dfki.de}\\
}

\begin{document}
\maketitle

\begin{abstract}
Decision trees can make vision-language model (VLM) classification interpretable by replacing a single prediction with a sequence of root-to-leaf questions. However, the accuracy cost of this greedy, non-backtracking procedure remains unclear. We introduce \textbf{MYOPIA}, a benchmark comprising four LLM-generated, human-refined decision trees for GTSRB, MNIST, CIFAR-10, and MSCOCO, together with matched prompts, diagnostic protocols, and an evaluation harness. Across GPT-4o, Qwen-VL Max, and LLaMA-3.2 11B Vision, tree traversal underperforms zero-shot prompting in all 12 dataset--model pairs; paired bootstrap analyses consistently favor zero-shot. Our diagnostics characterize interacting failure modes associated with this gap. Models often recall tree-encoded class attributes more accurately than they ground the same attributes in images: the mean label--image accuracy gaps are 9.5, 13.8, and 12.6 points for GPT-4o, Qwen-VL Max, and LLaMA, respectively. Incorrect local decisions then irreversibly exclude the correct class, while answer extraction introduces additional model-dependent loss. A node-level audit of the four trees finds that split entropy is weakly but consistently associated with lower node accuracy, whereas depth, branching factor, and within-node grounding differences are less stable. \textbf{MYOPIA} therefore provides a framework for diagnosing hierarchical visual classification and motivates testing tree designs that prioritize perceptual answerability and recovery from early errors. Code is available at \url{https://github.com/islammesabah/VLM_Classification}.
\end{abstract}
\section{Introduction}

Vision-language models (VLMs) can classify images without task-specific labeled examples. Pretrained on large image--text corpora, models such as CLIP, GPT-4o, and the Qwen-VL family perform strongly across a range of zero-shot visual tasks~\cite{radford2021learning, openai2024gpt4osystemcard, bai2023qwenvlversatilevisionlanguagemodel}. Nevertheless, two challenges remain. First, VLMs often struggle with fine-grained recognition that requires distinguishing subtle visual differences, such as traffic-sign variants, medical image subtypes, and car models~\cite{gan2025crossdomainmultistep, arora2025medfocusclip, wei2024enhancingfinegrainedimageclassifications}. This limitation is especially consequential in safety-critical domains. Second, opaque zero-shot predictions provide little insight into systematic errors~\cite{golovanevsky2025vlmsnotice}.

Decision trees offer a potential way to address both challenges. A tree decomposes a difficult classification decision into a sequence of simple, interpretable questions~\cite{nauta2021neuralprototypetrees, shi2019explainingpredictions}. Each answer narrows the label set until one class remains, and the resulting path explains the prediction. This rationale is intuitive, but it assumes that the model can answer every local question reliably. An early error excludes the correct class, prompt wording can alter intermediate answers, and longer paths create more opportunities for failure. Whether the added interpretability justifies these accuracy and inference costs has not been systematically evaluated.

We test this premise by comparing greedy, non-backtracking tree traversal with zero-shot prompting across three VLMs, four datasets, and several prompting variants. In the primary comparison, traversal underperforms zero-shot prompting in all 12 dataset--model pairs. We investigate this gap by separating class-attribute recall from visual grounding, distinguishing path-selection errors from answer-interface failures, and examining tree structure and inference cost. These findings do not show that hierarchical visual inference is inherently ineffective. Instead, they characterize the limitations of a common implementation and provide guidance for designing more reliable alternatives.

Our main contributions are:
\begin{enumerate}
    \item We introduce \textbf{MYOPIA}, a benchmark for evaluating tree-based reasoning in VLMs. It includes four LLM-generated, human-refined decision trees; matched zero-shot and tree-based prompt templates; paired class-attribute-recall and visual-grounding diagnostics; a reusable tree-generation pipeline (Section~\ref{sec:tree_structure}); and a complete evaluation harness.

    \item In the primary comparison, greedy tree traversal underperforms zero-shot prompting in every dataset--model pair, even when models can recall the relevant class attributes. Paired bootstrap analyses support this gap, and a GPT-4o pilot that adds the global label list to each tree prompt yields only limited and inconsistent gains (Section~\ref{sec:accuracy_comparison}). Thus, this implementation of greedy traversal does not reliably improve visual classification.

    \item We diagnose traversal failures by separating class-attribute recall from visual grounding (Section~\ref{sec:knowledge_verification}) and path-selection errors from answer-interface failures (Section~\ref{sec:traversal_error_analysis}). A node-level analysis examines associations between tree structure, grounding gaps, and node accuracy across the four evaluated trees (Section~\ref{sec:tree_quality_audit}), while prompt and cost analyses identify further design tradeoffs for interpretable VLM classifiers.
\end{enumerate}

\section{Related Work}

\textbf{Vision-Language Models and Zero-Shot Classification.}\\
VLMs such as CLIP~\cite{radford2021learning}, LLaVA~\cite{liu2023visualinstructiontuning}, the Qwen-VL family~\cite{bai2023qwenvlversatilevisionlanguagemodel, alibabacloud2026qwenvlmax}, and GPT-4o~\cite{openai2024gpt4osystemcard} have advanced zero-shot classification~\cite{li2025surveystateartlarge,CAO2025100069}, but performance often declines on fine-grained tasks with many semantically similar classes~\cite{gan2025crossdomainmultistep,arora2025medfocusclip,zhang2024visuallygroundedlanguagemodelsbad,kim-ji-2024-finer,wei2024enhancingfinegrainedimageclassifications}. Prompt learning~\cite{jha2024erapromptlearningvisionlanguage} and adaptation~\cite{Cho_2023_ICCV,DBLP:journals/corr/abs-2411-16407} can help, yet challenges remain in densely populated label spaces and with nuanced language~\cite{alhamoud2025visionlanguagemodelsunderstandnegation,anis2025limitationsvisionlanguagemodelsunderstanding,gou2024visionlanguagemodelsimage}.

\textbf{Multi-label and Hierarchical Classification.}
\label{sec:related_hierarchy}\\
Work on structured label spaces decomposes objects into attributes, states, and relationships~\cite{yellinek20253vlusingtreesimprove,ding2025treeattributespromptlearning}, classifies them across abstraction levels~\cite{zhang2024improvevisionlanguagemodel}, or augments classes with LLM-generated descriptions~\cite{saha2024improvedzeroshotclassificationadapting}. VLMs nevertheless remain sensitive to semantic granularity and textual specificity~\cite{xu2024benchmarkingzeroshotrecognitionvisionlanguage}.

Decision-tree classifiers provide path-based explanations~\cite{nauta2021neuralprototypetrees,shi2019explainingpredictions}, but require task-specific training or fitting rather than zero-shot VLM traversal. oVQA~\cite{ging2024openended} converts classification datasets into open-ended VQA and uses label hierarchies for follow-up questions; it evaluates answer granularity rather than greedy root-to-leaf classification.

Closest to our setting is CHiLS~\cite{novack2023chils}, which traverses an LLM-derived hierarchy from coarse to fine labels. Other approaches match images to LLM-generated descriptors~\cite{menon2023visual}, compare confusable classes using pairwise attributes~\cite{ren2023chatgpt}, or use GPT-4 descriptions as CLIP prompts~\cite{maniparambil2023enhancing}. These studies show that LLM-derived label structure can improve CLIP-style zero-shot classification. Unlike these embedding-based methods, our approach also structures inference as a sequence of VLM-answered branching questions.

\textbf{Structured Reasoning and Prompt Engineering.}\\
Structured reasoning has also been explored in language and visual tasks~\cite{besta2025demystifyingchainstreesgraphs,yellinek20253vlusingtreesimprove,wang2024videotree,he-etal-2024-advancing}. Tree of Thoughts~\cite{yao2023tree} and Graph of Thoughts~\cite{besta2024graph} allow models to explore multiple reasoning paths, backtrack, or merge candidates. On visual tasks, zero-shot chain-of-thought yields mixed results, helping larger models but hurting smaller ones~\cite{nagar2024zeroshot}. LLaVA-CoT~\cite{xu2025llavacot} and LlamaV-o1~\cite{thawakar2025llamavo1} use fine-tuning for multi-stage visual reasoning. We use zero-shot chain-of-thought as a free-form reasoning baseline without rigid branching (Section~\ref{sec:accuracy_comparison}).

\section{Tree-Based Classification Framework}

\begin{figure*}[t]
  \centering
  \includegraphics[width=0.75\linewidth]{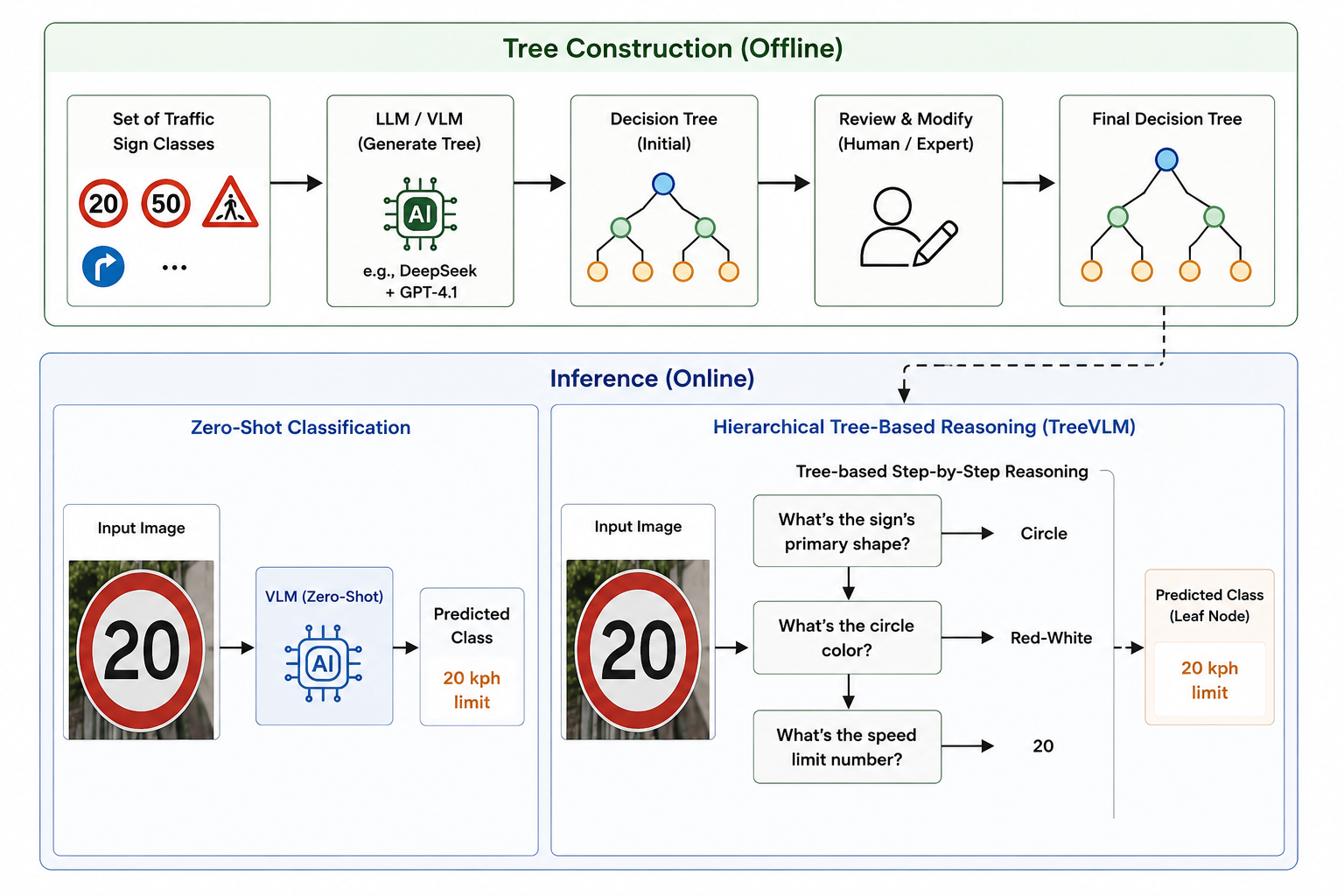}
  \caption{Overview of the experimental framework, showing how decision trees are constructed and how the same image is classified through direct zero-shot prompting or decision-tree traversal, with one question asked per node.}
  \label{fig:visual-abstract}
\end{figure*}

Our framework (Figure~\ref{fig:visual-abstract}) first constructs a decision tree that maps each class label to a distinct leaf, then uses a VLM to classify an image by traversing a root-to-leaf path.

\subsection{Tree Construction}
\label{sec:tree_structure}
An LLM (DeepSeek or GPT-4.1 in our experiments) generates each classification tree from class metadata and linguistic descriptions. The generation process avoids repeated questions along a path and recursively expands each branch until only one class remains, thereby guaranteeing exactly one class per leaf. A human annotator then traverses, reviews, and refines the tree to ensure semantic accuracy and consistency.

The procedure requires only a label set as input. At each internal node, a \emph{generator} proposes candidate questions, and an \emph{evaluator} scores them for clarity, completeness, visual relevance, logical grouping, and the balance of the resulting split. The highest-scoring question is selected, and the procedure recursively constructs its child branches. We release the implementation as a standalone script.

To assess structural transfer, we applied the procedure without modification to 20 bird species from CUB-200-2011~\cite{WahCUB_200_2011} not used elsewhere in this study. It produced a valid tree assigning all 20 species to distinct leaves, with a maximum depth of 6, 18 decision nodes (17 binary and one three-way), and 36 LLM calls. This result shows that the pipeline can construct a valid tree for a separate label set. Because we did not evaluate VLM classification on this tree, we make no accuracy claim.

\subsection{Tree Traversal}
\label{sec:method_implementation}
At inference, the model classifies an input image by traversing a single root-to-leaf path. At each internal node, the model answers the node's question, and its answer deterministically selects the next child. Each leaf maps to exactly one class label, such as \texttt{``20 kph speed limit''} or \texttt{``airplane''}. Each question is presented in a fresh, single-turn call containing the full image. Thus, traversing a path with $k$ internal nodes requires $k$ separate VLM calls.

\section{Benchmark and Experimental Setup}

MYOPIA benchmarks tree-based classification against zero-shot and chain-of-thought baselines across four datasets and three VLMs. Beyond end-to-end accuracy, it diagnoses three sources of performance loss: visual-grounding failure, path-selection error, and answer-interface failure.

\subsection{Datasets}
We evaluate four benchmark datasets spanning fine-grained discrimination and coarse-category classification:

\begin{itemize}
\item \textbf{GTSRB}~\cite{Stallkamp2012}: 43 fine-grained German traffic-sign classes from 50,000+ real-world images captured in multi-frame sequences. We sample one random frame per sequence, yielding \textbf{901} images (5--50 per class).

\item \textbf{MNIST}~\cite{lecun2010mnist}: 10 handwritten-digit classes requiring fine-detail discrimination despite visually simple images. We use \textbf{1,000 balanced samples} (100 per class).

\item \textbf{CIFAR-10}~\cite{krizhevsky2009learning}: 10 coarse-grained object classes testing generalization across broad categories. We use \textbf{1,000 balanced samples} (100 per class).

\item \textbf{MSCOCO}~\cite{lin2014microsoft}: a large-scale, coarse-grained recognition dataset comprising \textbf{91 object categories}. We construct the class hierarchy and zero-shot classification using \textbf{90 named object categories} available in the dataset. We select images containing exactly one of the \textbf{80 categories used in the standard object-detection benchmark} and sample up to 12 images per category, yielding an \textbf{850-image} set.
\end{itemize}

\subsection{Models}
Our benchmark evaluates three VLMs spanning different architectures, providers, and scales. \textbf{GPT-4o}~\cite{openai2024gpt4osystemcard}, accessed through OpenRouter with a fixed provider order, serves as the primary reference. \textbf{LLaMA-3.2 11B Vision Instruct}~\cite{meta2024llama32vision}, self-hosted at FP16 precision, offers a smaller, open-weight comparison. \textbf{Qwen-VL Max}~\cite{alibabacloud2026qwenvlmax} is accessed through Alibaba Cloud's DashScope API. Comparing its results with those of GPT-4o helps distinguish dataset-specific errors from model-specific errors.

\subsection{Prompting Strategies}
\label{sec:prompting}
We test three VLM strategies across all datasets and models, together with CLIP-based baselines on all datasets. \textbf{Zero-shot classification (baseline)}: the model receives the image with the prompt \texttt{"Classify the [object]. It should be only one of these classes: \{class\_ids\_and\_names\}. Respond with only the class ID."} A \textbf{with-descriptions} variant adds DeepSeek-generated per-class descriptions and a model-generated image caption.

\textbf{Chain-of-thought classification (CoT)}: to distinguish step-by-step reasoning from rigid, non-backtracking tree traversal, we include an additional baseline ($n=100$ class-stratified images per dataset-model pair) in which the model reasons step by step before returning its final class in a single call without a fixed tree.

\textbf{Tree-based classification}: the model traverses the constructed tree from root to leaf, each node prompted as \texttt{"\{tree.question\} Choose one of these answers: \{tree.answers\}."}; the chosen answer selects the next node. We test two variants: \textbf{with history}, appending prior decisions to each prompt, and \textbf{with descriptions}, adding the same class and image descriptions used in the zero-shot variant.

\textbf{CLIP-based baselines}: using frozen CLIP ViT-B/32~\cite{radford2021learning} without VLM calls, we evaluate plain zero-shot classification, description-based classification~\cite{menon2023visual}, and CHiLS~\cite{novack2023chils}. These methods rank classes by image--text embedding similarity rather than answering sequential branching questions. Plain zero-shot uses the prompt ``a photo of a \{class\_name\}''; description-based classification replaces each class name with its short description; and CHiLS reuses our hierarchies, treating each tree's root branches as superclasses.

\subsection{Benchmarking Sources of Performance Loss}
\label{sec:failure_benchmark}
We supplement end-to-end classification accuracy with a diagnostic benchmark targeting the three failure sources defined in our evaluation framework. All diagnostics use the same tree definitions and answer options as the classification experiments, allowing failures to be localized to visual grounding, branch selection, or the answer interface.

\textbf{Visual-grounding failure.} We compare matched label- and image-conditioned node-question protocols. In the label-conditioned condition, the model receives the ground-truth class label but no image and answers the questions associated with that class path while retaining conversation history. This measures \emph{class-attribute recall}: whether the model associates a class name with the attributes encoded by the tree. In the image-conditioned condition, the model receives the image but not the label and answers the same path questions. Image-conditioned probes use class-stratified subsamples of three to five images per class, with every class represented. We score each node answer and report per-class, per-depth, and aggregate accuracy. The label--image difference is treated as a diagnostic grounding gap rather than a causal estimate, because the two conditions provide different inputs.

\textbf{Path-selection error.} To determine where a traversal leaves the correct route, we reconstruct every stored path against the corresponding tree and ground-truth class. At each reached node, a parsed decision is correct if its branch still contains the ground-truth class; selecting any other branch is a path-selection error. Node- and depth-level accuracies are evaluated only along the ground-truth-consistent prefix and stop at the first path-selection or interface failure. Statistics at deeper nodes are therefore conditional on the traversal having reached them correctly.

\textbf{Answer-interface failure.}\label{sec:ans_ext} Because VLM output formats cannot be fully controlled, a strict regular-expression extractor returns the first predefined answer candidate found; if none is matched, the response and resulting path are marked unresolved. We report the unresolved-path rate and use a conservative oracle re-parser as a sensitivity analysis. Its rules cover recurring patterns identified through manual inspection, including number words, reordered or hyphenated labels, shape-adjective synonyms, and explicit affirmative or negated answers to yes/no questions; refusals and hedged responses remain unresolved. As an additional format control, we compare the free-form prompt with a numbered-choice format on matched GPT-4o subsamples ($n=20$ per dataset). Primary results retain the strict extractor and free-form prompt.

\subsection{Tree-Quality Audit Design}
\label{sec:tree_quality_audit_design}
Beyond the three failure diagnostics, we test whether tree structure is associated with node difficulty. Each internal node is identified by its root-to-node path. For node $v$ with $d_v$ children, let $C_{vj}$ be the classes reachable through child $j$ and $p_{vj}=|C_{vj}|/\sum_k|C_{vk}|$. We record depth, branching factor $d_v$, largest-child share $\max_jp_{vj}$, and split entropy $H_v=-\sum_jp_{vj}\log_2p_{vj}$. Entropy is reported in bits without normalization because branching factor is modeled separately.

The trees contain 128 internal nodes and 153 leaves (GTSRB 22/43, MNIST 8/10, CIFAR-10 9/10, and MSCOCO 89/90). GTSRB has 16 binary, two ternary, two five-way, and two eight-way nodes; MNIST has seven binary and one ternary node; and the other trees are binary. All internal nodes have at least two children. The audit retains 347 node--model pairs with both traversal accuracy and a grounding gap across 123 nodes (GPT-4o: 123; Qwen-VL Max: 122; LLaMA-3.2: 102); five MSCOCO nodes are never reached and are therefore excluded. We report bivariate correlations and compare evenly split nodes (largest-child share $\leq60\%$) with lopsided nodes.

We regress traversal accuracy on these structural variables, grounding gap, and model fixed effects. We report the baseline fit, node- and dataset-clustered standard errors, and a mixed-effects model with a node random intercept. Dataset-clustered inference is treated only as a sensitivity analysis because there are four clusters. A second sensitivity analysis weights observations by their numbers of scored decisions.

\subsection{Inference Cost}
We measure inference cost by counting model requests. Zero-shot and CoT classification require one VLM call per image, tree traversal requires one call per visited node, and CLIP-based methods require no VLM calls. We report the mean and maximum number of calls per image for each dataset.

\section{Evaluation and Results}

\subsection{Main Classification Results}
\label{sec:accuracy_comparison}

Because VLMs are sensitive to prompt phrasing~\cite{fatemi2023talklikegraphencoding}, we evaluated up to 10 zero-shot prompt variants at $T=0.7$ and report prompt~0 in Table~\ref{tab:zs_vs_tree}. Prompt sensitivity was substantial: prompt~0 was the strongest observed GTSRB variant for all three models, and the full-scale sweeps had standard deviations of 2.81--16.97 points (Supplementary Material). These evaluations vary prompt wording rather than repeat an unchanged prompt. Adding class and image descriptions had mixed effects: they improved LLaMA on CIFAR-10 by 17--19 points across temperatures and helped GPT-4o and Qwen-VL on MSCOCO, but reduced accuracy for all models on MNIST and for GPT-4o and LLaMA on GTSRB.

\begin{table}[ht]
\centering
\caption{Mean accuracy (\%) at $T=0.7$. ZS: zero-shot; D: class and image descriptions; CoT: chain-of-thought. Bold and underlined values denote the best and second-best results within each dataset--model row. CoT uses $n=100$ images per dataset--model pair.}
\footnotesize
\setlength{\tabcolsep}{5.8pt}
\begin{tabular}{@{}llrrrrr@{}}
\toprule
\textbf{Dataset} & \textbf{Model} & \textbf{ZS} & \textbf{ZS+D} & \textbf{CoT} & \textbf{Tree} & \textbf{Tree+D} \\
\midrule
GTSRB    & GPT-4o  & \textbf{71.14} & 63.60 & 63.00 & 57.71 & \underline{65.04} \\
         & LLaMA   & \textbf{37.07} & 23.53 & 12.00 & \underline{33.19}  & 14.87 \\
         & Qwen-VL & 47.06 & \underline{48.95} & 36.00 & 24.64  & \textbf{52.28} \\
\midrule
MNIST    & GPT-4o  & \underline{91.50} & 86.40 & \textbf{94.00} & 34.60  & 28.60 \\
         & LLaMA   & \underline{59.30} & 40.20 & \textbf{71.00} & 7.70 & 2.60 \\
         & Qwen-VL & \textbf{94.40} & \underline{89.20} & 86.00 & 29.30 & 27.10 \\
\midrule
CIFAR-10 & GPT-4o  & \underline{87.90} & \textbf{88.90} & 87.00 & 76.00 & 80.60 \\
         & LLaMA   & 36.70 & \textbf{54.00} & \underline{38.00} & 16.40 & 6.00 \\
         & Qwen-VL & \textbf{77.60} & \underline{70.90} & 63.00 & 56.60 & 66.30 \\
\midrule
MSCOCO   & GPT-4o  & \underline{72.82} & \textbf{80.82} & 71.00 & 46.71  & 44.00 \\
         & LLaMA   & \textbf{23.76} & \underline{21.06} & 18.00 & 3.65  & 0.35 \\
         & Qwen-VL & \underline{74.82} & \textbf{78.35} & 59.00 & 39.65 & 44.47 \\
\bottomrule
\end{tabular}
\label{tab:zs_vs_tree}
\end{table}

For tree-based classification at $T=0.7$, GPT-4o performs best on every dataset, whereas LLaMA performs worst overall. Descriptions again have model- and dataset-dependent effects and reduce accuracy for every model on MNIST. Adding the full decision history reduces accuracy in five of six GTSRB and CIFAR-10 comparisons; Qwen-VL on GTSRB is the only improvement (+12.3 points), while the losses range from 1.0 to 23.6 points.

\begin{table}[ht]
\centering
\caption{Instance-level Tree--ZS agreement at $T=0.7$, pooled across models. Both+/Both-- denote both correct/incorrect; Tree+/ZS+ denote only Tree/ZS correct.}
\footnotesize
\begin{tabular}{lrrrr}
\hline
\textbf{Dataset} & \textbf{Both+} & \textbf{Both-} & \textbf{Tree+} & \textbf{ZS+} \\ \hline
GTSRB    & 864  & 1127 & 177 & \textbf{535}  \\
MNIST    & 673  & 505  & 43  & \textbf{1779} \\
CIFAR-10 & 1333 & 821  & 157 & \textbf{689}  \\
MSCOCO   & 677  & 1005 & 88  & \textbf{780}  \\ \hline
\end{tabular}
\label{tab:instances_classification_comparison}
\end{table}

Greedy tree traversal underperforms zero-shot in all 12 dataset--model pairs (Table~\ref{tab:zs_vs_tree}). At the class level, pooled across the three models, zero-shot accuracy exceeded Tree accuracy in 72 of 129 class--model comparisons on GTSRB (55.8\%), 28 of 30 on MNIST (93.3\%), 27 of 30 on CIFAR-10 (90.0\%), and 182 of 240 on MSCOCO (75.8\%), using the 80 categories with sampled images. The instance-level results show the same pattern: zero-shot-only correct predictions outnumber tree-only correct predictions on every dataset (Table~\ref{tab:instances_classification_comparison}).

The same ordering holds on GTSRB and CIFAR-10 at $T=0$. GPT-4o and Qwen-VL are generally stable, although Qwen-VL's GTSRB tree accuracy rises from 24.64\% to 40.51\%. LLaMA is more sensitive: its GTSRB tree accuracy falls from 33.19\% to 6.33\%. Zero-shot nevertheless outperforms tree traversal for every model at both temperatures, so stochastic decoding does not explain the gap.

CoT outperforms tree traversal in 11 of 12 pairs, often substantially (MNIST/LLaMA: 71.0\% vs.\ 7.7\%), but exceeds zero-shot in only 3 of 12. It trails the tree only for GTSRB/LLaMA and trails zero-shot for all models on GTSRB and MSCOCO. This suggests that free-form reasoning is less damaging than rigid single-path branching but usually less effective than zero-shot, consistent with prior work~\cite{nagar2024zeroshot}. Because CoT and tree traversal also differ in call count, output format, and label access, the comparison does not isolate branching alone.

A paired bootstrap with 10,000 image resamples per pair confirms the gap: all 95\% confidence intervals favor zero-shot, with Tree--ZS deficits ranging from $-3.9$ to $-65.1$ points. All comparisons remain significant after Holm--Bonferroni correction (largest adjusted $p=0.028$). Adding the global class list to GPT-4o node prompts on pilot subsamples ($n=20$--$50$ per dataset) produces small, inconsistent changes ($-5$ to $+4$ points) and never approaches the corresponding zero-shot result, suggesting that label access is not the primary cause.

\subsection{Comparison to CLIP-Based Baselines}
\label{sec:clip_baselines}

To test whether hierarchical losses persist with embedding-based inference, we evaluate adapted frozen-CLIP baselines (Table~\ref{tab:clip_baselines}).

\begin{table}[ht]
\centering
\caption{Accuracy (\%) of CLIP ViT-B/32 baselines.}
\footnotesize
\setlength{\tabcolsep}{4pt}
\begin{tabular}{@{}lrrr@{}}
\toprule
\textbf{Dataset} & \textbf{CLIP ZS} & \textbf{Description} & \textbf{CHiLS} \\
\midrule
GTSRB    & 18.9 & 18.5 & \textbf{19.8} \\
MNIST    & 21.6 & \textbf{32.7} & 10.8 \\
CIFAR-10 & \textbf{89.0} & 86.5 & 79.0 \\
MSCOCO   & \textbf{69.6} & 66.7 & 66.9 \\
\bottomrule
\end{tabular}
\label{tab:clip_baselines}
\end{table}

Plain CLIP zero-shot is strongly dataset-dependent. Relative to the best VLM zero-shot result, it performs substantially worse on GTSRB and MNIST, slightly better on CIFAR-10, and 5.2 points worse on MSCOCO. Description-based classification improves MNIST from 21.6\% to 32.7\% but remains within three points of plain CLIP elsewhere. Unlike Menon and Vondrick~\cite{menon2023visual}, our implementation uses one sentence per class rather than multiple descriptors.

CHiLS improves plain CLIP by 0.9 points on GTSRB but reduces accuracy by 2.7 points on MSCOCO and about 10 points on CIFAR-10 and MNIST. The MNIST drop may partly reflect that the loop-count root split lacks a natural short caption. Thus, hierarchical CLIP scoring avoids the severe sequential-traversal loss on GTSRB and MSCOCO but remains harmful on the other datasets. Because CHiLS and VLM traversal differ in model, scoring rule, text representation, and hierarchy depth---CHiLS uses only root branches---this comparison does not isolate branching alone.

\subsection{Performance-Loss Benchmark Results}
\label{sec:knowledge_verification}
We next diagnose where tree-based classification loses accuracy by separating visual-grounding, path-selection, and answer-interface failures.

\textbf{Visual-grounding failure.} On GTSRB, GPT-4o achieves perfect label-conditioned path accuracy for 17 of 43 classes and 79.59\% macro-averaged accuracy over 231 node-question responses across the 43 class paths; shared internal nodes are evaluated once for each class whose path contains them. The six lowest-scoring classes, all at 33.3\%, are \textit{``End Truck Passing Ban''} and the five arrow-direction classes, whose names do not uniquely specify their visual form.

\begin{table}[ht]
\centering
\caption{Label-conditioned accuracy (L) and image-conditioned accuracy (I), with all classes represented.}
\footnotesize
\begin{tabular}{lrrrrrr}
\hline
& \multicolumn{2}{c}{\textbf{GPT-4o}} & \multicolumn{2}{c}{\textbf{LLaMA}} & \multicolumn{2}{c}{\textbf{Qwen-VL}} \\
\textbf{Dataset} & \textbf{L} & \textbf{I} & \textbf{L} & \textbf{I} & \textbf{L} & \textbf{I} \\ \hline
GTSRB    & 79.59 & 72.81 & 50.46 & 49.26 & 77.49 & 75.18 \\
MNIST    & 80.00 & 63.23 & 35.83 & 40.65 & 51.61 & 24.00 \\
CIFAR-10 & 97.50 & 89.71 & 100.00 & 57.71 & 100.00 & 82.29 \\
MSCOCO   & 95.79 & 89.33 & 87.74 & 76.09 & 94.73 & 87.14 \\ \hline
\textbf{Mean} & \textbf{88.22} & \textbf{78.77} & \textbf{68.51} & \textbf{55.93} & \textbf{80.96} & \textbf{67.15} \\ \hline
\end{tabular}
\label{tab:image_grounded}
\end{table}

For GPT-4o, replacing the label with the image reduces accuracy by 9.5 points on average (range 6.5--16.8; Table~\ref{tab:image_grounded}). MNIST has the largest gap, while GTSRB has one of the smallest despite also requiring fine-detail recognition, suggesting that grounding difficulty is dataset-specific rather than determined by class granularity alone. The grounding gap is also present across models, averaging 13.8 points for Qwen-VL (range 2.3--27.6) and 12.6 for LLaMA (range $-4.8$--42.3). On GTSRB, Qwen-VL reaches 75.18\%, close to GPT-4o's 72.81\%, whereas LLaMA reaches 49.26\%. Thus, visual-grounding difficulty is not specific to GPT-4o, although its magnitude varies substantially by model and dataset.

The GTSRB depth profile further localizes the gap (Table~\ref{tab:image_grounded_depth}). GPT-4o begins near ceiling at the root (97.7\%) but declines sharply from depth 2 onward. Qwen-VL dips at depths 1--2 before partly recovering, whereas LLaMA starts lower (55.3\%) and remains broadly flat.

\begin{table}[ht]
\centering
\caption{GTSRB image-conditioned accuracy by depth. Rows show depths 0--4; Overall includes all depths 0--14. GPT-4o and LLaMA use 215 images each, and Qwen-VL uses 129.}
\footnotesize
\begin{tabular}{lrrr}
\hline
\textbf{Depth} & \textbf{GPT-4o} & \textbf{LLaMA} & \textbf{Qwen-VL} \\ \hline
0 & 97.7\% & 55.3\% & 93.0\% \\
1 & 83.5\% & 47.5\% & 70.8\% \\
2 & 57.4\% & 51.8\% & 73.5\% \\
3 & 59.3\% & 53.3\% & 88.9\% \\
4 & 65.9\% & 42.4\% & 78.4\% \\
\textbf{Overall} & \textbf{72.81\%} & \textbf{49.26\%} & \textbf{75.18\%} \\ \hline
\end{tabular}
\label{tab:image_grounded_depth}
\end{table}

\textbf{Path-selection error.}\label{sec:traversal_error_analysis} Using the reconstruction protocol in Section~\ref{sec:failure_benchmark}, GTSRB node-level accuracies pooled across models are 86.7\%, 75.3\%, 76.3\%, 87.0\%, and 76.8\% at depths 0--4, respectively ($n=$2661/2002/1374/702/185). Accuracy varies non-monotonically with depth, indicating that errors are not confined to the deepest questions. Because each trace ends at its first incorrect or unresolved decision, the counts include only nodes reached along a ground-truth-consistent prefix.

\textbf{Answer-interface failure.} Under the strict extractor, unresolved-path rates range from 10.8\% on CIFAR-10 to 22.0\% on MSCOCO (Table~\ref{tab:inference_cost}). Oracle re-parsing reduces these rates on every dataset and raises pooled Tree accuracy from 38.51\% to 40.0\% on GTSRB, 23.9\% to 28.6\% on MNIST, 49.7\% to 53.5\% on CIFAR-10, and 30.0\% to 34.6\% on MSCOCO. These corrected values remain far below the corresponding zero-shot accuracies in Table~\ref{tab:zs_vs_tree}, showing that interface failures explain only part of the gap.

\begin{table}[ht]
\centering
\caption{Mean and maximum API calls per tree-classified image and unresolved-path rates before and after oracle re-parsing, pooled across models.}
\footnotesize
\begin{tabular}{lrrrr}
\hline
\textbf{Dataset} & \textbf{Mean calls} & \textbf{Max calls} & \textbf{Unresolved} & \textbf{Oracle} \\ \hline
GTSRB    & 3.61 & 15 & 14.7\% & 11.0\% \\
MNIST    & 2.45 & 4  & 18.6\% & 11.1\% \\
CIFAR-10 & 3.37 & 4  & 10.8\% & 3.7\% \\
MSCOCO   & 6.32 & 11 & 22.0\% & 16.2\% \\ \hline
\end{tabular}
\label{tab:inference_cost}
\end{table}

Interface failures are strongly model-specific. LLaMA accounts for 84.9\% and 97.0\% of unresolved paths on CIFAR-10 and MSCOCO, respectively, largely because it restates negated yes/no answers as full sentences. Its MSCOCO unresolved rate falls from 64.1\% to 47.5\% after oracle correction. LLaMA also accounts for 64.4\% of unresolved GTSRB paths, while MNIST is more evenly distributed (54.0\%). Thus, the answer interface imposes a real but uneven accuracy cost without overturning the central Tree--ZS comparison.

The numbered-choice control leads to little change for GPT-4o. Free-form responses are already nearly fully parseable on the matched \textbf{subsamples} (0.0--2.4\% unresolved), and node-level accuracy changes only modestly: 89.0\%$\to$90.7\% on GTSRB, 68.3\%$\to$71.4\% on MNIST, 92.3\%$\to$92.2\% on CIFAR-10, and 88.2\%$\to$86.7\% on MSCOCO. This suggests that GPT-4o's interface failures are concentrated in particular conditions rather than inherent to free-form responses.

\subsection{Tree-Quality Audit Results}
\label{sec:tree_quality_audit}
To identify which structural properties are associated with difficult node decisions, we first examine their bivariate relationships with traversal accuracy. Split entropy ($r=-0.24$) and branching factor ($r=-0.17$) correlate negatively with accuracy, whereas depth has a weak positive correlation ($r=0.18$). However, the depth correlation is negative within the CIFAR-10 ($-0.40$), MNIST ($-0.12$), and MSCOCO ($-0.08$) trees and positive only within GTSRB ($+0.56$), suggesting that the pooled result reflects between-tree differences. Evenly split nodes average 81.6\% accuracy ($n=68$), compared with 88.2\% for lopsided nodes ($n=55$).

We next analyze the 347 node--model observations using the multivariable specifications defined in Section~\ref{sec:tree_quality_audit_design}. The baseline fit has $R^2=0.16$, and with model fixed effects, LLaMA scores 15.5 points below GPT-4o ($95\%$ CI $[-21.1,-9.9]$, $p<0.0001$). Split entropy remains negative and significant in all four specifications ($\beta=-0.09$, $95\%$ CI $[-0.13,-0.04]$, $p\leq0.007$). Depth is significant in three ($\beta=+0.035$, $p=0.004$--$0.039$), and branching factor in two ($\beta=+0.052$, $p=0.029$--$0.11$); the latter changes sign after controlling for entropy, indicating collinearity. The grounding-gap coefficient is negative ($\beta=-0.03$) and significant in two specifications, but not in the node-random-intercept model ($p=0.23$). 

Thus, larger grounding gaps are associated with lower accuracy on average but not reliably within the same node, and the pooled model overstates this association ($\beta=-0.07$, $p<0.0001$). Split entropy is the only structural variable that remains negatively associated with accuracy after adjustment for the other factors, node dependence, and decision-count weighting. Because depth and branching factor lose significance under weighting, we treat split entropy as a weak correlate rather than a reliable predictor and limit this conclusion to the four evaluated trees.

\subsection{Inference Cost}
\begin{figure}[ht]
  \centering
  \includegraphics[width=\linewidth]{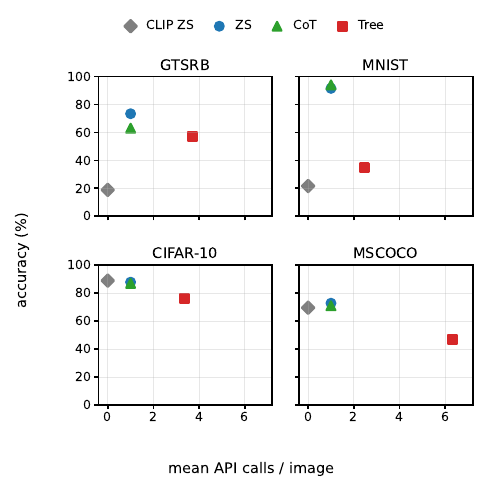}
  \caption{GPT-4o accuracy versus mean VLM calls per image at $T=0.7$; CLIP zero-shot requires no VLM calls. Tree traversal is dominated by zero-shot and CoT on every dataset.}
  \label{fig:cost_frontier}
\end{figure}

Tree traversal requires 2.5--6.3 times as many VLM calls per image as zero-shot classification (Table~\ref{tab:inference_cost}). MSCOCO requires the most calls because it has the deepest tree. Moreover, each node prompt resends the image, so every additional call incurs another image-token cost. For GPT-4o, tree traversal is dominated by zero-shot and CoT: it requires more calls and achieves lower accuracy on every dataset (Figure~\ref{fig:cost_frontier}). CLIP zero-shot requires no VLM calls and exceeds tree accuracy on CIFAR-10 and MSCOCO.

\section{Discussion}
\label{sec:discussion}

Taken together, the results challenge the assumption that an explicit hierarchy necessarily helps VLM classification. The comparisons with CoT and CLIP-based methods suggest that hierarchy alone is not the central problem. The observed deficits are consistent with irreversible local decisions as one contributor, but these comparisons do not isolate branching from differences in prompts, call counts, label access, or output formats. We therefore characterize the evaluated greedy implementation as a failure mode rather than establish rigid branching as a distinct causal source of degradation.

The diagnostic benchmark characterizes visual-grounding, path-selection, and answer-interface failures. Models can often recall tree-encoded attributes from a class name yet fail to recognize the same attributes in an image, particularly when nodes require fine perceptual distinctions. These local errors are not confined to deep levels, and an early mistake excludes the correct class. Parsing failures and conversation history add further, model-dependent loss, but neither accounts for the overall gap. Together, these patterns are consistent with an explanation involving perception, irreversible branching, and interface design rather than any single failure mode.

The tree-quality audit suggests a possible mismatch between conventional tree objectives and VLM behavior. Classical decision-tree induction commonly selects splits by maximizing information gain~\cite{quinlan1986induction}; in our deterministic class-partition setting, the split-entropy criterion rewards balanced branches. Across the four evaluated trees, higher split entropy is weakly but consistently associated with lower node accuracy. Because this observational relationship does not establish causation, it motivates testing whether perceptually answerable splits are preferable to balanced splits.

The cost analysis adds a practical concern: every additional node requires another image-conditioned call without an observed accuracy benefit in the evaluated settings. These results motivate testing hierarchies with fewer sequential decisions, coarse and visually grounded distinctions, constrained outputs, and alternatives retained through soft scoring or backtracking. More broadly, MYOPIA does not show that hierarchies are intrinsically ineffective; it provides a framework for separating knowledge, grounding, path-selection, interface, and cost failures and for studying when hierarchical classification breaks down.

\section{Limitations}
\label{sec:limitations}
\begin{itemize}
    \item \textbf{Scale and sampling.} Each evaluation set contains 850--1,000 images. MNIST and CIFAR-10 are balanced, MSCOCO is nearly balanced, and the 901-image GTSRB sample preserves the dataset's native imbalance. Larger evaluations may reveal additional failure modes.

    \item \textbf{Model coverage.} The three VLMs span different architectures and scales, but the findings may not generalize to other or future models.

    \item \textbf{Tree dependence.} The conclusions are tied to the evaluated trees. Although two human annotators and an independent LLM judge found no structural or ground-truth errors, the judge identified question-design concerns, particularly GTSRB's eight-way arrow question. Alternative or shallower trees may reduce the observed gap.

    \item \textbf{Tree-quality audit.} The audit is observational and limited to four fixed trees. Node accuracy is conditional on reach, coverage and decision counts vary across nodes, and dataset-clustered inference uses only four clusters. The reported associations should therefore not be interpreted causally or generalized beyond these trees.

    \item \textbf{Extraction and robustness.} The primary evaluation relies on regex answer extraction; oracle re-parsing provides only a sensitivity analysis (Section~\ref{sec:ans_ext}). Run-to-run variation was evaluated at full scale only for GTSRB and CIFAR-10. Smaller MNIST and MSCOCO checks remained stable (standard deviation $\leq2.9$ points), but do not replace full-scale replication.

    \item \textbf{Cost measurement.} We report API-call counts rather than exact monetary costs. 

\end{itemize}

\section{Conclusion}

Sequential, single-path tree traversal did not improve VLM classification in the evaluated settings. Across four datasets, zero-shot prompting outperformed the tree for the three models, with paired bootstrap confidence intervals supporting the gap (Section~\ref{sec:accuracy_comparison}). This finding does not rule out hierarchical inference; instead, it shows that greedy, non-backtracking traversal underperforms a simpler baseline and motivates controlled alternatives.

The diagnostic benchmark identifies multiple failure modes associated with the gap rather than establishing a single cause. Models often recall class attributes but cannot ground them reliably in images; an incorrect local decision then removes the correct class, while answer extraction adds further model-dependent loss (Section~\ref{sec:knowledge_verification}). In the four evaluated trees, the tree-quality audit finds a weak association between higher split entropy and lower node accuracy; it does not establish that balanced splits cause difficult visual decisions (Section~\ref{sec:tree_quality_audit}). These findings motivate testing tree-construction objectives that prioritize perceptual answerability, robust interfaces, and recovery from early errors.

\paragraph{Future work.}
Future studies should isolate branching through controlled comparisons that hold the model, prompt, label access, and inference budget fixed. Promising directions include multi-branch or soft routing, confidence-based backtracking, constrained decoding, and joint optimization of tree structure for visual answerability and cost. Evaluating multiple independently generated trees, additional VLMs, and larger or multi-object datasets would further test generalizability.

\paragraph{Reproducibility.}
\label{sec:reproducibility}
The supplementary material reports complete $T=0$, tree-with-history, 10-prompt zero-shot, global-label, numbered-choice, and run-to-run results, along with full tree-audit specifications and node counts. We will release the trees, prompts, generator--evaluator pipeline, analysis code, and evaluation harness under the MIT License; dataset access remains governed by the original licenses (Section~\ref{sec:ethics}).
\section{Ethics and Broader Impact}
\label{sec:ethics}

This work uses only public benchmark datasets (GTSRB~\cite{Stallkamp2012}, MNIST~\cite{lecun2010mnist}, CIFAR-10~\cite{krizhevsky2009learning}, MSCOCO~\cite{lin2014microsoft}), collects no new human-subject data, and introduces no new capability, so we see no dual-use risk; images may incidentally contain identifiable people or vehicles. \textbf{Licensing.} GTSRB and CIFAR-10 require citation, MNIST is CC BY-SA 3.0, MSCOCO annotations are CC BY 4.0; we redistribute only annotations and sample-ID lists, and release the trees for research, not deployment.

{
  \small
  \bibliographystyle{ieeenat_fullname}
  \bibliography{main}
}

\def\supplementinpaper{}

\ifdefined\supplementinpaper
\clearpage
\begin{center}
{\Large\bfseries MYOPIA: Understanding the Limits of Tree-Based Reasoning\\
in Vision-Language Models\\[0.4em]Supplementary Material\par}
\end{center}
\vspace{-1em}
\else
\documentclass[10pt,twocolumn,letterpaper]{article}
\usepackage[margin=0.75in]{geometry}
\usepackage{booktabs}
\usepackage{graphicx}
\usepackage{amsmath}
\usepackage{amssymb}
\usepackage{times}
\usepackage[hidelinks]{hyperref}
\fi
\setcounter{table}{0}
\renewcommand{\thetable}{S\arabic{table}}
\setcounter{figure}{0}
\renewcommand{\thefigure}{S\arabic{figure}}
\setcounter{section}{0}
\renewcommand{\thesection}{S\arabic{section}}
\renewcommand{\theHtable}{supp.\arabic{table}}
\renewcommand{\theHfigure}{supp.\arabic{figure}}
\renewcommand{\theHsection}{supp.\arabic{section}}

\ifdefined\supplementinpaper
\else
\title{\vspace{-2em}MYOPIA: Understanding the Limits of Tree-Based Reasoning\\
in Vision-Language Models\\[0.4em]\large Supplementary Material}
\author{}
\date{}

\begin{document}
\raggedbottom
\maketitle
\vspace{-3em}
\fi

\section{Scope and Conventions}
\label{sec:supp_scope}

This supplement provides the complete results referenced in the main paper's
\textbf{Reproducibility paragraph}:
\begin{itemize}
    \item Section~\ref{sec:supp_full} reports results for all available
    prompting conditions at $T{=}0$ and $T{=}0.7$, including the
    tree-with-history ablation.
    \item Section~\ref{sec:supp_stats} reports paired-bootstrap intervals and
    Holm-corrected tests for the tree--zero-shot gap.
    \item Section~\ref{sec:supp_prompts} evaluates ten paraphrases of the
zero-shot prompt, quantifies the resulting spread, and locates the prompt
reported in the main paper within that spread.
\item Section~\ref{sec:supp_repeats} explains that \texttt{RunId} indexes
distinct prompt wordings, not repeated calls with the same prompt. 
    \item Sections~\ref{sec:supp_global} and~\ref{sec:supp_numbered} report the
    global-label and numbered-choice probes, respectively.
    \item Sections~\ref{sec:supp_trees}--\ref{sec:supp_audit} document the tree
    specifications and audit coverage.
    \item Section~\ref{sec:supp_repro} describes the released artifacts, fixed
    image subsets, decoding settings, and limits on exact re-execution.
\end{itemize}

The following conventions apply to all reported results.

\paragraph{Accuracy denominator.} A prediction is correct only when its parsed
integer class identifier matches the ground-truth identifier. The denominator
includes every row in a condition; unparseable responses and images without a
logged traversal are counted as errors. Reported end-to-end accuracy therefore
incorporates answer-interface failures as well as classification failures.
Interface effects are analyzed separately in the main paper and in
Section~\ref{sec:supp_numbered}.

\paragraph{Deduplication.} Rows in
{\urlstyle{same}\nolinkurl{all_sequences_training/}} duplicate current GTSRB
images in {\urlstyle{same}\nolinkurl{datasets/GTSRB/Training/}}. All tables use
a view that removes a legacy row when a matching current row exists for the
same model, temperature, ablation, and run. Without this step, GTSRB
denominators increase from 901 to as many as 915 and accuracies shift by up to
0.3 points.

\paragraph{Coverage.} A full pass contains 901 GTSRB images, 1,000 MNIST
images, 1,000 CIFAR-10 images, and 850 MSCOCO images. This
supplement reports every available cell and marks incomplete cells with their
$n$.

\section{Complete Results at Both Temperatures}
\label{sec:supp_full}

Table~\ref{tab:supp_full} reports six prompting conditions at $T{=}0$ and
$T{=}0.7$, extending Table~1 of the main paper with the recorded $T{=}0$
results and the tree-with-history (Tree+H) ablation.

At $T{=}0$, only GTSRB and CIFAR-10 have complete passes. The available MNIST
and MSCOCO cells contain 2--860 images and should not be interpreted as
dataset-level estimates; they are included with their $n$ for transparency.
Accordingly, the main paper discusses $T{=}0$ only for GTSRB and CIFAR-10.

Tree+H likewise has complete coverage only for GTSRB and CIFAR-10. At
$T{=}0.7$, it reduces accuracy in five of the six dataset--model comparisons;
Qwen-VL on GTSRB is the sole exception. The MNIST and MSCOCO Tree+H cells
contain only 2--33 images and are reported as incomplete observations.

Boldface and underlining identify the best and second-best eligible entries in
each row. Full-pass conditions and the planned $n{=}100$ CoT probes are
eligible; superscripted incomplete cells are excluded.

\begin{table*}[t]
\centering
\caption{Accuracy (\%) across temperatures and prompting conditions. ZS: zero-shot; +D: class and image descriptions; +H: conversation history across tree nodes; CoT: chain-of-thought.  Superscripts give $n$ for incomplete cells, and an em dash indicates no data. CoT uses $n{=}100$ class-stratified images per dataset--model pair at $T{=}0.7$. Boldface and underlining identify the best and second-best entries among full-pass conditions and the planned CoT probes; superscripted cells are excluded. Full-pass sizes are 901 (GTSRB), 1,000 (MNIST and CIFAR-10), and 850 (MSCOCO).}
\label{tab:supp_full}
\footnotesize
\setlength{\tabcolsep}{4.6pt}
\begin{tabular}{@{}ll c rrrrrr@{}}
\toprule
\textbf{Dataset} & \textbf{Model} & $T$ & \textbf{ZS} & \textbf{ZS+D} & \textbf{CoT} & \textbf{Tree} & \textbf{Tree+H} & \textbf{Tree+D} \\
\midrule
GTSRB     & GPT-4o   & $0$ &   \textbf{70.92} &            63.82 &              --- &            58.05 &            55.83 & \underline{66.81} \\
          &          & $0.7$ &   \textbf{71.14} &            63.60 &            63.00 &            57.71 &            56.71 & \underline{65.04} \\
\cmidrule(l){2-9}
          & LLaMA    & $0$ &   \textbf{23.86} & \underline{18.42} &              --- &             6.33 &            16.09 &            15.65 \\
          &          & $0.7$ &   \textbf{37.07} &            23.53 &            12.00 & \underline{33.19} &            12.43 &            14.87 \\
\cmidrule(l){2-9}
          & Qwen-VL  & $0$ &            47.95 & \underline{49.72} &              --- &            40.51 &            36.96 &   \textbf{51.83} \\
          &          & $0.7$ &            47.06 & \underline{48.95} &            36.00 &            24.64 &            36.96 &   \textbf{52.28} \\
\midrule
MNIST     & GPT-4o   & $0$ &  94.87$^{\,234}$ &  88.02$^{\,860}$ &              --- &  35.47$^{\,234}$ &   46.67$^{\,15}$ &              --- \\
          &          & $0.7$ & \underline{91.50} &            86.40 &   \textbf{94.00} &            34.60 &   75.00$^{\,16}$ &            28.60 \\
\cmidrule(l){2-9}
          & LLaMA    & $0$ &    33.33$^{\,3}$ &              --- &              --- &    33.33$^{\,3}$ &     0.00$^{\,2}$ &              --- \\
          &          & $0.7$ & \underline{59.30} &            40.20 &   \textbf{71.00} &             7.70 &    50.00$^{\,2}$ &             2.60 \\
\cmidrule(l){2-9}
          & Qwen-VL  & $0$ &  95.17$^{\,414}$ &              --- &              --- &  14.25$^{\,414}$ &              --- &              --- \\
          &          & $0.7$ &   \textbf{94.40} & \underline{89.20} &            86.00 &            29.30 &    7.69$^{\,13}$ &            27.10 \\
\midrule
CIFAR-10  & GPT-4o   & $0$ & \underline{89.60} &   \textbf{90.00} &              --- &            75.40 &            73.10 &            82.70 \\
          &          & $0.7$ & \underline{87.90} &   \textbf{88.90} &            87.00 &            76.00 &            72.80 &            80.60 \\
\cmidrule(l){2-9}
          & LLaMA    & $0$ & \underline{43.70} &   \textbf{62.80} &              --- &            17.10 &            12.30 &             4.80 \\
          &          & $0.7$ &            36.70 &   \textbf{54.00} & \underline{38.00} &            16.40 &            14.10 &             6.00 \\
\cmidrule(l){2-9}
          & Qwen-VL  & $0$ &   \textbf{77.60} & \underline{71.20} &              --- &            56.60 &            33.00 &            66.50 \\
          &          & $0.7$ &   \textbf{77.60} & \underline{70.90} &            63.00 &            56.60 &            33.00 &            66.30 \\
\midrule
MSCOCO    & GPT-4o   & $0$ &   73.91$^{\,92}$ &              --- &              --- &   46.74$^{\,92}$ &     0.00$^{\,7}$ &              --- \\
          &          & $0.7$ & \underline{72.82} &   \textbf{80.82} &            71.00 &            46.71 &   33.33$^{\,33}$ &            44.00 \\
\cmidrule(l){2-9}
          & LLaMA    & $0$ &    50.00$^{\,2}$ &              --- &              --- &     0.00$^{\,2}$ &     0.00$^{\,4}$ &              --- \\
          &          & $0.7$ &   \textbf{23.76} & \underline{21.06} &            18.00 &             3.65 &    33.33$^{\,6}$ &             0.35 \\
\cmidrule(l){2-9}
          & Qwen-VL  & $0$ &  76.47$^{\,221}$ &              --- &              --- &  40.27$^{\,221}$ &              --- &              --- \\
          &          & $0.7$ & \underline{74.82} &   \textbf{78.35} &            59.00 &            39.65 &              --- &            44.47 \\
\bottomrule
\end{tabular}
\end{table*}

\section{Significance of the Tree--Zero-Shot Gap}
\label{sec:supp_stats}

Table~\ref{tab:supp_full} shows the tree condition below zero-shot almost
everywhere. Table~\ref{tab:supp_bootstrap} attaches an interval and a
corrected $p$-value to that gap, so the claim rests on more than a point
estimate.

The design is paired: for a given dataset, model, and temperature, the
zero-shot and tree predictions for one image are stored on the same database
row, so the two conditions see identical images. The bootstrap therefore
resamples \emph{images}, not predictions, and recomputes both accuracies on
each resample; the reported interval is over the difference, which absorbs the
correlation between the two conditions rather than treating them as
independent samples.

Two limits are worth stating. With 10,000 resamples, the two-sided bootstrap
$p$ cannot be resolved below $2/10{,}000$. Eleven comparisons reach this
floor, so their corrected values are reported as upper bounds rather than
exact quantities. The intervals quantify sampling variation over images at a
fixed decoding temperature.

All twelve gaps are negative, and all twelve remain significant after
Holm--Bonferroni correction over the family of twelve comparisons (largest
adjusted $p=0.0278$, GTSRB with LLaMA). The single smallest gap,
$-3.88$ points for LLaMA on GTSRB, is also the only one whose interval
approaches zero ($[-7.33, -0.44]$); every other interval excludes zero by a
wide margin.

\begin{table*}[t]
\centering
\caption{Paired bootstrap over images for the tree-minus-zero-shot accuracy gap at $T{=}0.7$ (no descriptions, no history, run~0). Images are resampled with replacement ($10{,}000$ resamples; seed~42), and both accuracies are recomputed on each resample. $\Delta$ is Tree$-$ZS in percentage points; the interval is the 2.5th--97.5th percentile range of the resampled $\Delta$. $p_{\text{Holm}}$ is the two-sided bootstrap $p$ after Holm--Bonferroni correction across the twelve dataset--model comparisons. Values preceded by $<$ are upper bounds imposed by bootstrap resolution.}
\label{tab:supp_bootstrap}
\footnotesize
\setlength{\tabcolsep}{5pt}
\begin{tabular}{@{}llrrrrcr@{}}
\toprule
\textbf{Dataset} & \textbf{Model} & $n$ & \textbf{ZS} & \textbf{Tree} & $\Delta$ & \textbf{95\% CI} & $p_{\text{Holm}}$ \\
\midrule
GTSRB & GPT-4o & 901 & 71.14 & 57.71 & $-13.43$ & $[-16.43, -10.54]$ & $<$0.0024 \\
GTSRB & LLaMA & 901 & 37.07 & 33.19 & $-3.88$ & $[-7.33, -0.44]$ & 0.0278 \\
GTSRB & Qwen-VL & 901 & 47.06 & 24.64 & $-22.42$ & $[-25.64, -19.20]$ & $<$0.0024 \\
\midrule
MNIST & GPT-4o & 1000 & 91.50 & 34.60 & $-56.90$ & $[-60.00, -53.70]$ & $<$0.0024 \\
MNIST & LLaMA & 1000 & 59.30 & 7.70 & $-51.60$ & $[-55.00, -48.10]$ & $<$0.0024 \\
MNIST & Qwen-VL & 1000 & 94.40 & 29.30 & $-65.10$ & $[-68.10, -62.10]$ & $<$0.0024 \\
\midrule
CIFAR-10 & GPT-4o & 1000 & 87.90 & 76.00 & $-11.90$ & $[-14.50, -9.40]$ & $<$0.0024 \\
CIFAR-10 & LLaMA & 1000 & 36.70 & 16.40 & $-20.30$ & $[-23.80, -16.80]$ & $<$0.0024 \\
CIFAR-10 & Qwen-VL & 1000 & 77.60 & 56.60 & $-21.00$ & $[-24.10, -17.90]$ & $<$0.0024 \\
\midrule
MSCOCO & GPT-4o & 850 & 72.82 & 46.71 & $-26.12$ & $[-29.88, -22.47]$ & $<$0.0024 \\
MSCOCO & LLaMA & 850 & 23.76 & 3.65 & $-20.12$ & $[-23.06, -17.18]$ & $<$0.0024 \\
MSCOCO & Qwen-VL & 850 & 74.82 & 39.65 & $-35.18$ & $[-38.82, -31.41]$ & $<$0.0024 \\
\bottomrule
\end{tabular}
\end{table*}

\section{Ten-Prompt Zero-Shot Sweep}
\label{sec:supp_prompts}

Section~5.1 of the main paper evaluates ten zero-shot prompts at $T{=}0.7$ and
reports one of them in Table~1. This section gives the full sweep behind that
choice. Ten paraphrases of the zero-shot instruction were written per dataset;
they differ in register and verbosity but request the same output, a single
class ID\@. The prompt index is stored as \texttt{RunId}, and prompt~0 is the
one reported in main-paper Table~1.

\begin{figure}[t]
\centering
\includegraphics[width=\linewidth]{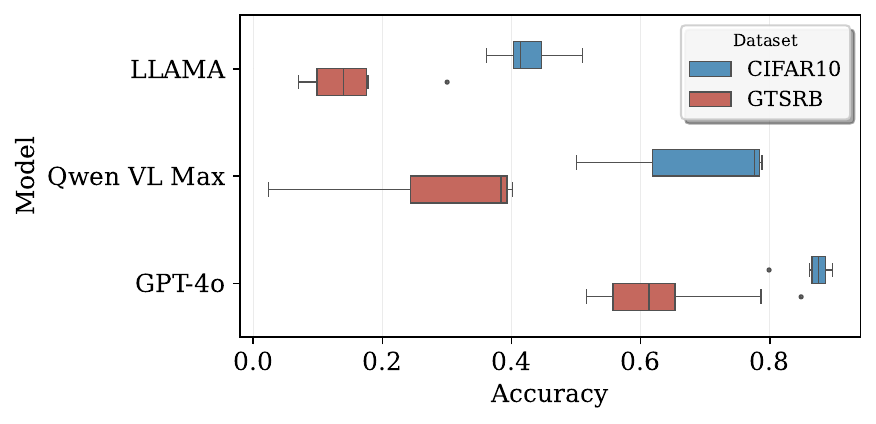}
\caption{Prompt sensitivity of zero-shot accuracy at $T{=}0.7$. Dots show
full-coverage prompt results, diamonds mark prompt~0 (reported in main-paper
Table~1), black ticks show sweep means, and horizontal lines span the observed
range. Red denotes GTSRB and blue denotes CIFAR-10. On GTSRB, prompt~0 is the
highest full-coverage prompt for all three models; the GPT-4o/GTSRB row spans
eight prompts rather than ten.}
\label{fig:supp_prompt_sweep}
\end{figure}

Table~\ref{tab:supp_prompt_sweep} and Figure~\ref{fig:supp_prompt_sweep} give
per-prompt accuracy for the six dataset--model pairs swept at full scale. Five
pairs contain ten full-coverage prompts; GPT-4o on GTSRB contains eight,
because P2 and P8 stopped at $n{=}78$ and $n{=}108$ respectively. Summary
statistics exclude those two cells, and the partial accuracies are not
comparable to the rest of the row.

\paragraph{Prompt sensitivity is a first-order effect.} Across-prompt standard
deviations range from 2.81 points (CIFAR-10, GPT-4o) to 16.97 points (GTSRB,
Qwen-VL), and the max--min spread reaches 41.5 points on GTSRB for Qwen-VL and
31.1 points for LLaMA\@. Prompt choice is therefore comparable in magnitude to
the differences between models, and any single zero-shot number should be read
as one draw from this spread. Part of the spread is discontinuous rather than
gradual: Qwen-VL falls to 5.55\% on one GTSRB paraphrase and to 50.10\% on one
CIFAR-10 paraphrase, drops more consistent with format non-compliance than with
a loss of recognition ability.

\paragraph{Where prompt~0 sits in the distribution.} On CIFAR-10, prompt~0 is
close to the sweep mean for GPT-4o (87.90\% versus 87.08\%), below it for LLaMA
(36.70\% versus 42.13\%), and above it for Qwen-VL (77.60\% versus 71.01\%). On
GTSRB it is the highest full-coverage prompt for all three models: 71.14\%,
37.07\%, and 47.06\%, against sweep means of 59.67\%, 16.57\%, and 35.84\%.
The GTSRB zero-shot column of main-paper Table~1 is therefore drawn from the
favorable end of its prompt distribution, while the CIFAR-10 column is
representative.

\paragraph{Effect on the Tree--zero-shot comparison.} Tree traversal was
evaluated once, with a fixed node prompt that does not vary with
\texttt{RunId} (Section~\ref{sec:supp_repeats}), so the sweep bears on only one
side of the comparison. On GTSRB, mean zero-shot accuracy stays above tree
accuracy for GPT-4o (59.67\% versus 57.71\%) and Qwen-VL (35.84\% versus
24.64\%), but falls below it for LLaMA (16.57\% versus 33.19\%); several
individual prompts also reverse the GPT-4o and Qwen-VL gaps. On CIFAR-10 the
zero-shot mean stays above tree accuracy for all three models. The twelve
negative gaps in Table~\ref{tab:supp_bootstrap} thus describe the prompt~0
comparison specifically. The direction of the result is robust on CIFAR-10 and
for the aggregate, but on GTSRB its magnitude, and for LLaMA its sign, depend
on zero-shot prompt wording.

\begin{table*}[t]
\centering
\caption{Zero-shot accuracy (\%) for each of the ten prompt paraphrases at $T{=}0.7$, on the six dataset--model pairs swept at full scale. P0, underlined, is the prompt reported in main-paper Table~1. Em dashes mark runs that did not reach full coverage and are excluded from the summary statistics. Mean, SD, and $\Delta$ are computed over the available prompts, with $\Delta=\max-\min$.}
\label{tab:supp_prompt_sweep}
\footnotesize
\setlength{\tabcolsep}{4pt}
\begin{tabular}{@{}llrrrrrrrrrr@{\hspace{6pt}}rrr@{}}
\toprule
\textbf{Dataset} & \textbf{Model} & \textbf{P0} & \textbf{P1} & \textbf{P2} & \textbf{P3} & \textbf{P4} & \textbf{P5} & \textbf{P6} & \textbf{P7} & \textbf{P8} & \textbf{P9} & \textbf{Mean} & \textbf{SD} & $\boldsymbol{\Delta}$ \\
\midrule
GTSRB     & GPT-4o   & \underline{71.14} &            63.71 &              --- &            66.04 &            45.84 &            56.16 &            67.70 &            49.94 &              --- &            56.83 & 59.67 & 8.94 & 25.31 \\
          & LLaMA    & \underline{37.07} &            23.20 &            15.54 &            20.75 &            11.10 &            17.09 &            18.87 &             8.10 &             5.99 &             7.99 & 16.57 & 9.28 & 31.08 \\
          & Qwen-VL  & \underline{47.06} &            46.95 &            46.06 &            45.84 &             5.55 &            45.84 &            46.06 &            11.88 &            17.31 &            45.84 & 35.84 & 16.97 & 41.51 \\
\midrule
CIFAR-10  & GPT-4o   & \underline{87.90} &            86.20 &            88.70 &            86.50 &            89.70 &            89.50 &            86.70 &            87.20 &            79.90 &            88.50 & 87.08 & 2.81 &  9.80 \\
          & LLaMA    & \underline{36.70} &            51.00 &            43.10 &            36.10 &            41.20 &            41.60 &            40.90 &            45.10 &            40.10 &            45.50 & 42.13 & 4.38 & 14.90 \\
          & Qwen-VL  & \underline{77.60} &            78.40 &            76.50 &            50.10 &            77.70 &            57.00 &            78.50 &            56.90 &            78.60 &            78.80 & 71.01 & 11.45 & 28.70 \\
\bottomrule
\end{tabular}
\end{table*}

\section{Scope of Repeated Measurement}
\label{sec:supp_repeats}

The repeated-measurement field in the released databases does not index
independent stochastic replications. \texttt{RunId} selects the zero-shot
prompt paraphrase, and all ten indices correspond to distinct prompt texts.
The variation reported in Section~\ref{sec:supp_prompts} therefore measures
sensitivity to wording, not variation across repeated calls with an unchanged
prompt.

Tree traversal was executed under \texttt{RunId}~0 alone. Its prompt is built
from the current node's question and answer set and does not vary with the
zero-shot prompt index, so the database contains exactly one tree traversal per
dataset--model--temperature--ablation cell. Reported tree accuracies are
single-run measurements. Replicated tree runs remain an important robustness
check.

Where a full-scale sweep is unavailable, Table~\ref{tab:supp_matched_prompts}
reports the available prompt checks on MNIST and MSCOCO, restricted to the
images shared by all prompts so that the spread reflects wording rather than
differing image subsets. Three of the four checks have SD $\leq 2.9$ points;
GPT-4o on MSCOCO has SD 4.79 points over a 65.0--75.0\% range. Each check uses
20 images, so these estimates are imprecise and do not replace a full-scale
sweep.

\begin{table}[t]
\centering
\caption{Small-scale zero-shot prompt checks on the datasets without a full-scale sweep. Each row compares four prompts on the same matched image subset, so the spread is not confounded by which images each prompt saw. Mean, SD, and $\Delta$ are percentages, with $\Delta=\max-\min$.}
\label{tab:supp_matched_prompts}
\footnotesize
\begin{tabular}{@{}llrrrrr@{}}
\toprule
\textbf{Dataset} & \textbf{Model} & $n$ & \textbf{Prompts} & \textbf{Mean} & \textbf{SD} & $\boldsymbol{\Delta}$ \\
\midrule
MNIST     & GPT-4o   & 20 & 4 & 93.75 & 2.50 &  5.00 \\
          & Qwen-VL  & 20 & 4 & 97.50 & 2.89 &  5.00 \\
MSCOCO    & GPT-4o   & 20 & 4 & 68.75 & 4.79 & 10.00 \\
          & Qwen-VL  & 20 & 4 & 68.75 & 2.50 &  5.00 \\
\bottomrule
\end{tabular}
\end{table}

\section{Global-Label Probe}
\label{sec:supp_global}

Tree traversal shows the model only the current node's answer options, whereas
zero-shot prompting shows the complete label set. If restricted label access
caused most of the Tree--zero-shot gap, appending all labels to each node prompt
should recover a substantial part of it. We test this hypothesis with GPT-4o
on matched, class-stratified subsamples of 20--50 images per dataset. Each image
is evaluated once with the local options and once with the global label list
appended.

Table~\ref{tab:supp_global} reports both conditions on their shared images. In
dataset order (GTSRB, MNIST, CIFAR-10, MSCOCO), the global list changes node
accuracy by $+1.0$, $-5.3$, $-1.4$, and $+0.1$ points; image accuracy changes
by $+4$, $0$, $-5$, and $+4$ points. The unresolved rate also increases on
three of the four datasets, consistent with the longer prompt making format
compliance more difficult in some cases.

The probe therefore provides no evidence that label access is the primary
cause of the gap: it produces no large or consistent recovery toward the
full-run zero-shot results. Because the matched samples are small, however,
the probe cannot resolve effects of only a few points and should not be read as
evidence of exactly zero effect.

\begin{table}[t]
\centering
\caption{Global-label probe with GPT-4o on images shared by both conditions. Local uses only the current node's answer options; Global also appends the dataset's full class list. Node accuracy and unresolved rate are computed over visited node decisions, and image accuracy is end-to-end traversal accuracy. All values are percentages.}
\label{tab:supp_global}
\footnotesize
\setlength{\tabcolsep}{4pt}
\begin{tabular}{@{}lrrrrrrr@{}}
\toprule
& & \multicolumn{2}{c}{\textbf{Node acc.}} & \multicolumn{2}{c}{\textbf{Unresolved}} & \multicolumn{2}{c}{\textbf{Image acc.}} \\
\cmidrule(lr){3-4} \cmidrule(lr){5-6} \cmidrule(l){7-8}
\textbf{Dataset} & $n_{\text{img}}$ & \textbf{Local} & \textbf{Global} & \textbf{Local} & \textbf{Global} & \textbf{Local} & \textbf{Global} \\
\midrule
GTSRB     & 50 & 89.1 & 90.1 & 2.1 & 3.1 & 58.0 & 62.0 \\
MNIST     & 20 & 71.1 & 65.8 & 0.0 & 0.0 & 35.0 & 35.0 \\
CIFAR-10  & 20 & 88.9 & 87.5 & 1.6 & 7.8 & 65.0 & 60.0 \\
MSCOCO    & 50 & 91.2 & 91.3 & 0.3 & 3.8 & 46.0 & 50.0 \\
\bottomrule
\end{tabular}
\end{table}

\begin{table}[!t]
\centering
\caption{Numbered-choice probe with GPT-4o ($n{=}20$ images per dataset). Free uses strict extraction from a free-form response; Num.\ requests only an option number. Node accuracy and unresolved rate use each condition's visited nodes; image accuracy is end-to-end. All values are percentages.}
\label{tab:supp_numbered}
\footnotesize
\setlength{\tabcolsep}{4pt}
\begin{tabular}{@{}lrrrrrrr@{}}
\toprule
& & \multicolumn{2}{c}{\textbf{Node acc.}} & \multicolumn{2}{c}{\textbf{Unresolved}} & \multicolumn{2}{c}{\textbf{Image acc.}} \\
\cmidrule(lr){3-4} \cmidrule(lr){5-6} \cmidrule(l){7-8}
\textbf{Dataset} & $n_{\text{img}}$ & \textbf{Free} & \textbf{Num.} & \textbf{Free} & \textbf{Num.} & \textbf{Free} & \textbf{Num.} \\
\midrule
GTSRB     & 20 & 89.0 & 90.7 & 1.4 & 0.0 & 60.0 & 65.0 \\
MNIST     & 20 & 68.3 & 71.4 & 2.4 & 0.0 & 35.0 & 30.0 \\
CIFAR-10  & 20 & 92.3 & 92.2 & 1.5 & 0.0 & 75.0 & 75.0 \\
MSCOCO    & 20 & 88.2 & 86.7 & 0.0 & 0.0 & 35.0 & 30.0 \\
\bottomrule
\end{tabular}
\end{table}

\begin{table}[!t]
\centering
\caption{Numbered-choice probe on nodes reached at the same image and depth in both conditions. $\Delta$ is numbered minus free-form node accuracy, in percentage points.}
\label{tab:supp_numbered_paired}
\footnotesize
\begin{tabular}{@{}lrrrr@{}}
\toprule
\textbf{Dataset} & \textbf{Paired nodes} & \textbf{Free} & \textbf{Num.} & $\boldsymbol{\Delta}$ \\
\midrule
GTSRB     & 73 & 89.0 & 90.4 & +1.4 \\
MNIST     & 36 & 69.4 & 77.8 & +8.3 \\
CIFAR-10  & 63 & 92.1 & 92.1 & +0.0 \\
MSCOCO    & 105 & 90.5 & 86.7 & -3.8 \\
\bottomrule
\end{tabular}
\end{table}

\section{Numbered-Choice Probe}
\label{sec:supp_numbered}

Because the strict extractor marks a response unresolved whenever it fails to
match a predefined answer, part of the tree deficit could be an
answer-interface artifact rather than a reasoning failure. This probe controls
for that possibility by asking each node twice on the same 20 images per
dataset: once with the free-form prompt and strict regular-expression
extraction, and once with a numbered-option prompt that requests only an option
number.

Table~\ref{tab:supp_numbered} reports both conditions. The numbered format
eliminates unresolved responses (0.0\% on all four datasets), while free-form
responses are already nearly fully parseable for GPT-4o (0.0--2.4\%
unresolved). The corresponding node-accuracy changes are $+1.7$, $+3.1$,
$-0.1$, and $-1.5$ points for GTSRB, MNIST, CIFAR-10, and MSCOCO,
respectively. Removing unresolved responses therefore does not produce a
consistent accuracy improvement.

\paragraph{A caveat on matching.} Each traversal terminates at its first
incorrect or unresolved decision, so the two conditions do not visit the same
number of nodes even on the same image: a condition that fails earlier
accumulates fewer, shallower, and therefore easier nodes. The node accuracies
in Table~\ref{tab:supp_numbered} are computed over each condition's own node
set, which is what the main paper reports.
Table~\ref{tab:supp_numbered_paired} instead restricts the analysis to nodes
reached at the same depth on the same image in both conditions. Matching in
this stricter sense increases the MNIST gain from $+3.1$ to $+8.3$ points and
deepens the MSCOCO loss from $-1.5$ to $-3.8$ points, while GTSRB and CIFAR-10
are essentially unchanged.

\paragraph{A caveat on power.} The probe uses 20 images per dataset, yielding
36--110 node decisions per condition. None of the four exploratory differences
is distinguishable from zero (Fisher exact, all $p>0.75$), and the image-level
direction reverses on two datasets. These tests also treat node decisions as
independent and do not model their nesting within images. The probe therefore
rules out only a large interface effect for GPT-4o; it does not establish
equivalence between the formats. Notably, the free-form unresolved rates are
well below GPT-4o's 9.9\% rate in the full GTSRB run, indicating that its
interface failures are concentrated in particular conditions rather than
inherent to free-form prompting.

\section{Tree Specifications}
\label{sec:supp_trees}

Table~\ref{tab:supp_tree_specs} summarizes the four evaluated trees. Every
internal node has at least two children, every leaf maps to exactly one class,
and no question repeats along a root-to-leaf path. In the branching column,
for example, $16\times2$ denotes 16 binary internal nodes.

For an internal node $v$, let $p_{vj}$ be the fraction of descendant leaves
assigned to child $j$. We define split entropy as
$H_v=-\sum_j p_{vj}\log_2 p_{vj}$. Higher values indicate a more even division
of the remaining classes. The values are not normalized by the maximum entropy
for the node's branching factor.

The trees differ substantially in shape. MNIST and CIFAR-10 are shallow
(maximum depth 3) with 8--9 internal nodes. MSCOCO is a binary tree with 89
internal nodes and maximum depth 10. GTSRB is the most irregular: it combines
16 binary nodes with two ternary, two five-way, and two eight-way nodes and
reaches depth 14. An independent LLM judge flagged the eight-way GTSRB
arrow-direction question as a possible question-design concern; it is also the
highest-entropy split in the four trees (3.00 bits).

Five MSCOCO internal nodes were never reached by a primary-condition
traversal. Because they have no measured traversal accuracy, they are excluded
from the audit, leaving 123 of 128 internal nodes with operational estimates.

\begin{table*}[t]
\centering
\caption{Specifications of the four evaluated trees. Int.\ is the number of internal nodes; Depth is the maximum root-to-leaf depth. Branching gives internal-node outdegrees and their multiplicities. Split entropy is reported in unnormalized bits. Unreached counts internal nodes that no primary-condition traversal visited.}
\label{tab:supp_tree_specs}
\footnotesize
\setlength{\tabcolsep}{4pt}
\begin{tabular}{@{}lrrrlrrr@{}}
\toprule
\textbf{Dataset} & \textbf{Int.} & \textbf{Leaves} & \textbf{Depth} & \textbf{Branching} & $\boldsymbol{\bar{H}}$ & $\boldsymbol{H}$ \textbf{range} & \textbf{Unreached} \\
\midrule
GTSRB     & 22 & 43 & 14 & 16$\times$2, 2$\times$3, 2$\times$5, 2$\times$8 & 1.02 & 0.35--3.00 & 0 \\
MNIST     & 8 & 10 & 3 & 7$\times$2, 1$\times$3 & 1.01 & 0.92--1.30 & 0 \\
CIFAR-10  & 9 & 10 & 3 & 9$\times$2 & 0.96 & 0.81--1.00 & 0 \\
MSCOCO    & 89 & 90 & 10 & 89$\times$2 & 0.93 & 0.54--1.00 & 5 \\
\midrule
\textbf{Total} & \textbf{128} & \textbf{153} & & & & & \textbf{5} \\
\bottomrule
\end{tabular}
\end{table*}

\section{Tree-Quality Audit Coverage}
\label{sec:supp_audit}

Table~\ref{tab:supp_audit_coverage} shows how the 347 retained node--model
observations are distributed across datasets and models. The audit covers 123
distinct internal nodes, yielding 369 possible node--model combinations. A
combination is retained only when both conditional in-traversal accuracy and a
grounding-gap estimate are available; 22 combinations are therefore excluded.

Coverage is uneven. MSCOCO contributes 237 of the 347 observations (68\%)
from 84 nodes, whereas MNIST contributes 22 observations from 8 nodes. Any
pooled association is therefore weighted toward MSCOCO, the only deep binary
tree in the set. The audit's conclusions are accordingly limited to these four
evaluated trees rather than presented as general properties of hierarchical
classifiers.

Per-model coverage is 123 observations for GPT-4o, 122 for Qwen-VL Max, and
102 for LLaMA-3.2. Because traversal stops after the first failure, node
accuracy is conditional on reach: deeper nodes are scored only on traversals
that completed all earlier decisions correctly. It should therefore be read as
an operational quantity, not as an intrinsic measure of node answerability.

\begin{table}[t]
\centering
\caption{Tree-quality audit coverage. Dataset rows report distinct retained nodes and node--model pairs; model rows report retained pairs. Each pair requires both conditional in-traversal accuracy and a grounding gap.}
\label{tab:supp_audit_coverage}
\footnotesize
\begin{tabular}{@{}lrr@{}}
\toprule
\textbf{Dataset} & \textbf{Nodes} & \textbf{Node--model pairs} \\
\midrule
\multicolumn{3}{@{}l}{\textit{By dataset}} \\
GTSRB     & 22 & 61 \\
MNIST     & 8 & 22 \\
CIFAR-10  & 9 & 27 \\
MSCOCO    & 84 & 237 \\
\midrule
\multicolumn{3}{@{}l}{\textit{By model}} \\
GPT-4o & & 123 \\
LLaMA & & 102 \\
Qwen-VL & & 122 \\
\midrule
\textbf{Total} & \textbf{123} & \textbf{347} \\
\bottomrule
\end{tabular}
\end{table}

\section{Reproducibility Details}
\label{sec:supp_repro}

This section records what a reader needs to regenerate the reported results and
which parts of the evaluation cannot be reproduced exactly.

\paragraph{Artifacts in this supplement.} The accompanying archive contains
the four evaluated decision trees in both human-readable and machine-readable
forms; the complete prompt set, including zero-shot paraphrases, per-class
descriptions, and node-question templates; the generator--evaluator
tree-construction pipeline; the evaluation harness; and the analysis scripts
that produce every table in this document. It excludes API credentials, raw
provider traces, and the datasets themselves.

\paragraph{Sampling.} Each dataset is evaluated on a fixed image subset listed
by identifier in the archive: 901 GTSRB, 1,000 MNIST, 1,000 CIFAR-10, and 850
MSCOCO images. The MSCOCO subset is restricted to images with exactly one
category label, ensuring that each image has a single ground-truth class. The
identifier lists constitute the sampling record; the subsets are not redrawn
at run time.

\paragraph{Decoding.} All primary conditions use temperature $0$ or $0.7$, one
image per request, no few-shot examples, and the endpoint or deployment
defaults for other decoding parameters. Tree traversal issues one request per
visited node. The tree-with-history ablation additionally carries previous node
questions and answers forward in the conversation.

\paragraph{Determinism and its limits.} The analysis layer is deterministic:
the bootstrap uses seed~42 and 10,000 resamples, and every table is recomputed
from the stored databases rather than from cached summaries. Model inference
is not guaranteed to be bit-reproducible, even at $T{=}0$. GPT-4o and Qwen-VL
Max are accessed through hosted services whose implementations or checkpoints
may change. LLaMA-3.2 11B Vision Instruct is self-hosted from fixed official
weights at FP16 precision, but hardware and software nondeterminism can still
affect its outputs. A new evaluation run may preserve the qualitative patterns
without reproducing the exact percentages.

\paragraph{Recomputation path.} The stored databases are the authoritative
source for every table reported here. Given these files, the archived analysis
scripts regenerate all tables and the figure without model access or API
credentials. Reproducing the databases themselves requires access to the
hosted services and the self-hosted LLaMA deployment, with the associated API
and compute costs.

\paragraph{Scoring requirements.} Exact recomputation requires the GTSRB
deduplication described in Section~\ref{sec:supp_scope}; otherwise, denominators
increase from 901 to as many as 915. Accuracy must also count unparseable
answers and unlogged traversals as errors. Restricting evaluation to parseable
responses raises the tree results and makes them incomparable with the values
reported here.

\ifdefined\supplementinpaper
\def\endsupplement{\clearpage}
\else
\def\endsupplement{\end{document}}
\fi
\endsupplement

\end{document}